\documentclass[preprint,12pt]{elsarticle}

\usepackage{lineno,hyperref}
\modulolinenumbers[5]

\graphicspath{ {./fig/} }
\usepackage[a4paper, total={6in, 9in}]{geometry}
\usepackage{latexsym}
\usepackage{graphicx}
\usepackage{amsfonts,amssymb,amsmath}
\usepackage{caption}
\usepackage{booktabs}
\usepackage{textcomp}
\usepackage{gensymb}
\usepackage{bm}
\usepackage{url}
\usepackage{multirow}
\usepackage{multicol}
\usepackage{lineno}
\usepackage{setspace}
\usepackage{xcolor}
\usepackage{subcaption}
\usepackage{makecell}
\usepackage{ragged2e}
\usepackage{array}
\usepackage{draftwatermark}

\usepackage[ruled,vlined]{algorithm2e}
\usepackage{etoolbox}
\newcommand{\algorithmicdoinparallel}{\textbf{do in parallel}}
\makeatletter
\AtBeginEnvironment{algorithm2e}{%
  \newcommand{\FORALLP}[2][default]{\ALC@it\algorithmicforall\ #2\ %
    \algorithmicdoinparallel\ALC@com{#1}\begin{ALC@for}}%
}
\makeatother

\journal{Journal of \LaTeX\ Templates}

\journal{arXiv}

\begin{document}

\SetWatermarkScale{5}
%\SetWatermarkAngle{30}
\SetWatermarkText{Preprint}

\begin{frontmatter}

\title{Fully Data-Driven Model for Increasing Sampling Rate Frequency of Seismic Data using Super-Resolution Generative Adversarial Networks}

 \author[ua]{Navid Gholizadeh}
\author[ub]{Javad Katebi\corref{cor1}}

\address[ua]{Department of Civil and Environmental Engineering, Amirkabir University of Technology, Tehran, Iran}
\address[ub]{Faculty of Civil Engineering, University of Tabriz, Tabriz, Iran}
 \cortext[cor1]{Corresponding author}


%\onehalfspacing
\begin{abstract}
High-quality data is one of the key requirements for any engineering application. In earthquake engineering practice, accurate data is pivotal in predicting the response of structure or damage detection process in an Structural Health Monitoring (SHM) application with less uncertainty. However, obtaining high-resolution data is fraught with challenges, such as significant costs, extensive data channels, and substantial storage requirements. To address these challenges, this study employs super-resolution generative adversarial networks (SRGANs) to improve the resolution of time-history data such as the data obtained by a sensor network in an SHM application, marking the first application of SRGANs in earthquake engineering domain. The time-series data are transformed into RGB values, converting raw data into images. SRGANs are then utilized to upscale these low-resolution images, thereby enhancing the overall sensor resolution. This methodology not only offers potential reductions in data storage requirements but also simplifies the sensor network, which could result in lower installation and maintenance costs. The proposed SRGAN method is rigorously evaluated using real seismic data, and its performance is compared with traditional enhancement techniques. The findings of this study pave the way for cost-effective and efficient improvements in the resolution of sensors used in SHM systems, with promising implications for the safety and sustainability of infrastructures worldwide.
\end{abstract}


\end{frontmatter}

%\linenumbers
%\doublespacing

\section{Introduction}
In a structural engineering practice, data is not only required in analysis and design phase for better response prediction but also it is needed in Structural Health Monitoring (SHM) phase for accurate and timely model updating, system identification, and damage detection and finally in reliability assessment and risk-based decision-making. Quality of the input data is one of the uncertainty resources. In the case of time-series data, better quality refers to both 1) accurate quantity of desired parameters in a specific time and 2) smaller time steps.

Usually there is a trade-off between accuracy, cost, and simplicity. Quality of the data can be affected in first place during data collection process due to inaccurate and low-frequency data accusation systems or later in application process due to simplified methods such as Fourier or modified inverse Fourier transforms to estimate a complex time-history function \cite{Faroughi}.

%In this method, signals that have low amplitude, thus less participation in overall time-history, are neglected resulting in higher uncertainty to a degree. 

However, there are cases that need to use an accurate time-history with smaller time step \cite{Phillips} which requires high-resolution data or high-frequency sensor. This is mainly beneficial for identification of more deterioration and collapse modes or simply for uncertainty reduction purposes. Moreover, achieving convergence in nonlinear analysis typically involves using smaller increments in loading \cite{Mei}. In the context of time-history analysis, this is accomplished by employing smaller time steps. In scenarios involving low-frequency data and a reduced analysis time step, the midpoints of the input time-history are determined through linear interpolation. This approach contributes to an increase in uncertainty.

As our built environment grows and ages, the demand for sophisticated monitoring systems that continuously collect data to assess the structural condition of structural systems becomes imperative. Central to the effectiveness of structural analysis and SHM methods is the resolution of the data, which directly impacts the accuracy and timeliness of damage prediction and detection \cite{resol1}. In addition, the importance of sampling-rate in structural health monitoring of bridges is emphasized in \cite{YU201760}.

The modern approach towards ensuring the structural integrity of buildings, especially tall ones, in areas with high seismic activity has evolved to be performance-based, mandating the installation of seismic instrumentation for real-time structural health monitoring. This instrumentation generates data that is crucial for model updating, system identification, and damage detection. Guidelines and design codes like those from the Los Angeles Tall Buildings Structural Design Council (LATBSDC) \cite{naeim2008} and Pacific Earthquake Engineering Research Center's Tall Buildings Initiative (PEER TBI) \cite{Pacific} embody this modern approach.

Several research studies have emphasized the significance of high-resolution data in SHM. To this end, the role of sensor resolution in early-stage crack detection was highlighted in \cite{resol2} and it was demonstrated in \cite{resol3} that a self-powered broadband vibration sensor capable of detecting high-frequency vibrations ranging from 3 to 133 kHz and offering excellent frequency resolution, can identify even minor frequency changes, making it suitable for identifying minor defects in applications such as SHM. The significance of minimum sampling rate for reliable fatigue lifetime estimation is discussed in \cite{Pietro}. 

However, obtaining high-resolution data comes with several challenges. Conventional sensors, such as accelerometers, strain gauges, string potentiometers, and LVDTs, yield data points at discrete intervals \cite{challenge1, challenge3}. On the other hand, high-resolution sensors, while offering superior data quality, come at a considerable cost, demand an extensive array of data channels for transmission, and necessitate significant data storage capacity \cite{challenge2}. The need for installation of hundreds of these types of sensors and the needed amount of time for these data to be preserved (usually 5-10 years based on structural provisions or contract between consultant engineer and owner) even results in higher cost. Wind and acceleration data from the Hardanger bridge \cite{data} is an example for comparison purpose, where high frequency data almost occupies 890 GB in respect to 17 GB space of low frequency data. The importance of sampling rate and consumed computational resources is further discussed in \cite{Tang}.

A few studies have explored methods to improve sensor resolution in SHM. Traditional techniques, such as interpolation \cite{Interpolation} and signal processing \cite{signal_processing}, have been employed to enhance data quality. While these methods yield notable improvements, they often face limitations in terms of computational complexity and the preservation of details.

As new architectures in data science emerge, they are used in different engineering practices to reduce the amount of uncertainty in variety of applications. Super-resolution generative adversarial networks (SRGAN) is a cutting-edge deep learning model used for upscaling low-resolution images to high-resolution ones. By leveraging adversarial training, SRGANs generate visually impressive results, making them valuable in fields like medical imaging \cite{GAN_medical1, GAN_medical2}, satellite image enhancement \cite{GAN_sattelite}, and document restoration \cite{recovery}. In this study, SRGAN is employed for the first time to enhance the resolution of earthquake engineering data such as data captured by sensors in SHM applications. The sensor data is transformed into RGB values, effectively converting raw data into images. Subsequently, SRGAN is utilized to elevate the resolution of these images, thereby increasing the overall sensor resolution. For SHM, the benefits of SRGANs are multifaceted. Firstly, there's a significant reduction in data storage requirements. Additionally, SRGANs can potentially simplify the sensor network, reducing installation and maintenance costs. In general, the main contributions of this study are as follows.

\begin{itemize}
\item {Employing SRGAN to increase the resolution of earthquake engineering data for the first time;}
\item {Rigorous evaluation of the developed model using real seismic data \cite{PEERDatabase, PEER201303} consisting ground motion acceleration, velocity, and displacement measurements;}
\item {Comparison of the proposed SRGAN method with other traditional methods.}
\end{itemize}

The remainder of this paper is organized as follows. Section \ref{sys_model} describes data preparation and SRGAN-based methodology. Section \ref{results} details experimental procedures and outcomes. Section \ref{Conclusion} offers conclusions and future research directions.

\section{System Model}\label{sys_model}
In this section, a comprehensive overview of data preprocessing process and SRGAN-based system model is provided. Additionally, the section highlights the evaluation metrics utilized to assess the performance of the SRGAN-based system. 

\subsection{Data Preprocessing}
\begin{figure}[h!]
	\begin{center}
		\includegraphics[width=0.6\columnwidth]{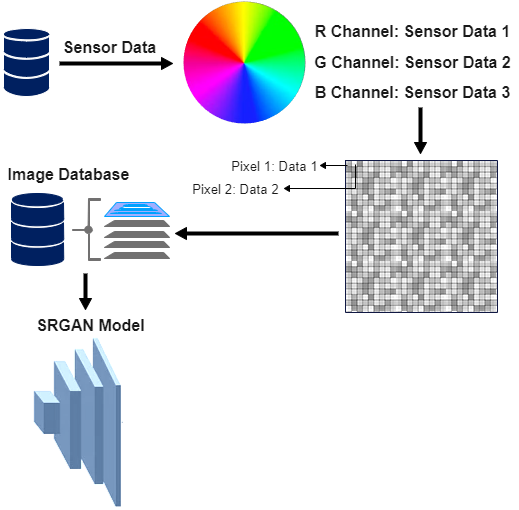}
		\caption{System Architecture}\label{arch}
	\end{center}
\end{figure}

In this study, the processed sensor data obtained from the PEER Ground Motion Database \cite{PEERDatabase} is analyzed. Both horizontal components for ground-motion set proposed by FEMA P695 guidelines \cite{fema} are selected for training and testing purpose in this study. The Record Sequence Numbers (RSNs) for these data is provided in Table \ref{t1}. The dataset comprises ground motion acceleration, velocity, and displacement data collected through specialized sensors. In this data preprocessing procedure, sensor measurements from the dataset were transformed into a format suitable for image analysis. The original dataset comprised three columns, each representing a distinct sensor measurement. To leverage the potential of image processing techniques, the numerical values were transformed into 136x136 pixel images. Each sensor measurement was interpreted as the value for one of the RGB channels, and the sensor measurements were normalized to fit within the RGB range of 0 to 255. This transformation allowed for the visualization and analysis of the sensor data in a manner akin to interpreting images, enabling the application of various computer vision algorithms. A schematic of this approach is illustrated in Fig. \ref{arch}.

\begin{table}[h!]
 	\centering
 	\caption{FEMA P695 Ground-Motion Set}
  \tabcolsep=0.4cm
 	\begin{tabular}{ccc}
 	\hline
 	 	\multicolumn{3}{c}{\textbf{Record Sequence Number (RSN)}}\\
 		\hline
 		  Far-Field & Near-Field (Pulse) & Near-Field (No Pulse)\\
            \hline
            68      &     181     &     126     \\
            125     &     182     &     160     \\
            169     &     292     &     165     \\
            174     &     $723^{*}$    &     495     \\
            721     &     802     &     $496^{*}$    \\
            $725^{*}$    &     821     &     741     \\
            752     &     828     &     753     \\
            767     &     879     &     825     \\
            $829^{**}$   &     1063    &     1004    \\
            848     &     1086    &     $1048^{*}$   \\
            900     &     1165    &     1176    \\
            953     &     1503    &     1504    \\
            960     &     1529    &     1517    \\
            1111    &     1605    &     2114    \\
            1116    &             &             \\
            1148    &             &             \\
            1158    &             &             \\
            1244    &             &             \\
            1485    &             &             \\
            1602    &             &             \\
            1633    &             &             \\
            1787    &             &             \\
            \hline
            \multicolumn{3}{l}{\footnotesize{* Vertical component for these ground-motions isn’t available.}}\\
            \multicolumn{3}{l}{\makecell[l]{\footnotesize{** Currently, these ground-motions aren’t available through PEER}\\ \footnotesize{~~~~Ground Motion Database.}}}\\
 		\hline
 	\end{tabular}%
 	\label{t1}%
 \end{table}%

\subsection{SRGAN Architecture}
SRGAN is an advanced machine learning model designed for image super-resolution tasks. The primary objective is to reconstruct high-resolution (HR) images from their low-resolution (LR) counterparts. SRGAN leverages the power of Generative Adversarial Networks (GANs) to achieve this and consists of two main components: a generator and a discriminator. The generator aims to upscale LR images to HR images. It often employs a deep convolutional neural network architecture, which uses a series of convolutional, ReLU, and batch normalization layers. The goal is to map the feature space of LR images to that of HR images in a way that retains or even adds detail, making the upscaled image visually similar to an actual HR image. The discriminator is a type of neural network designed to differentiate genuine HR images from artificial ones produced by the generator. Essentially, it operates as a binary classifier, trained to recognize real HR images with high probability while assigning a low probability to the generated ones.

During the training phase, the generator and discriminator engage in a sort of game. The generator tries to produce HR images so convincing that the discriminator can't distinguish them from real HR images. Conversely, the discriminator aims to become better at distinguishing real from fake. The two networks are trained simultaneously through this adversarial process.

An LR image is passed through the generator to produce a synthetic HR image. The discriminator evaluates the synthetic HR image against a real HR image. Two types of losses are often used. One is the content loss, calculated using features from a pre-trained network (like VGG19) to ensure that the generated and real HR images are semantically similar. The other is the adversarial loss, aimed at making the generated HR image indistinguishable from real HR images in the eyes of the discriminator. Gradients are computed and both the generator and discriminator are updated accordingly. The end result is a generator capable of upscaling LR images with high fidelity, producing results that are often indistinguishable from real HR images.

The content loss is often calculated using the mean squared error (MSE) between feature representations of the generated and real HR images with pixel dimensions of $W \times H$. These feature representations are generally obtained from a pre-trained network like VGG19. Let \( I_{\text{HR}} \) be the real HR image and \( G(I_{\text{LR}}) \) be the generated HR image from the LR input \( I_{\text{LR}} \). Let \( \phi \) be the feature extractor function. The content loss \( \mathcal{L}_{\text{content}} \) is given by:

\begin{equation}
\mathcal{L}_{\text{content}} = \frac{1}{W \times H} \sum_{x=1}^{W} \sum_{y=1}^{H} \left[ \phi(I_{\text{HR}})_{xy} - \phi(G(I_{\text{LR}}))_{xy} \right]^2
\label{e1}
\end{equation}

Let \( D \) be the discriminator network. Then, the adversarial loss \( \mathcal{L}_{\text{adv}} \) is given by:

\begin{equation}
\mathcal{L}_{\text{adv}} = -\log(D(G(I_{\text{LR}})))
\label{e2}
\end{equation}

Since the pixel-wise difference is very important in this study, an additional penalty term is also added to the generator's loss. This loss, called pixel loss, represents the MSE loss between the generated and real HR image pixels. The final loss \( \mathcal{L}_{\text{total}} \) for the generator is a weighted sum of the content, adversarial, and pixel losses:

\begin{equation}
\mathcal{L}_{\text{total}} = \mathcal{L}_{\text{content}} + \lambda \mathcal{L}_{\text{adv}} + \beta \mathcal{L}_{\text{pixel}}
\label{e3}
\end{equation}

Here, \( \lambda \) is a hyperparameter that controls the trade-off between the content and adversarial losses. For the discriminator, the loss is usually a binary cross-entropy loss calculated on both real and generated HR images.

\begin{equation}
\mathcal{L}_{\text{D}} = - \left[ \log(D(I_{\text{HR}})) + \log(1 - D(G(I_{\text{LR}}))) \right]
\label{e4}
\end{equation}

\begin{figure}[h!]
    \centering
    \begin{subfigure}[b]{1\textwidth}
        \includegraphics[width=\textwidth]{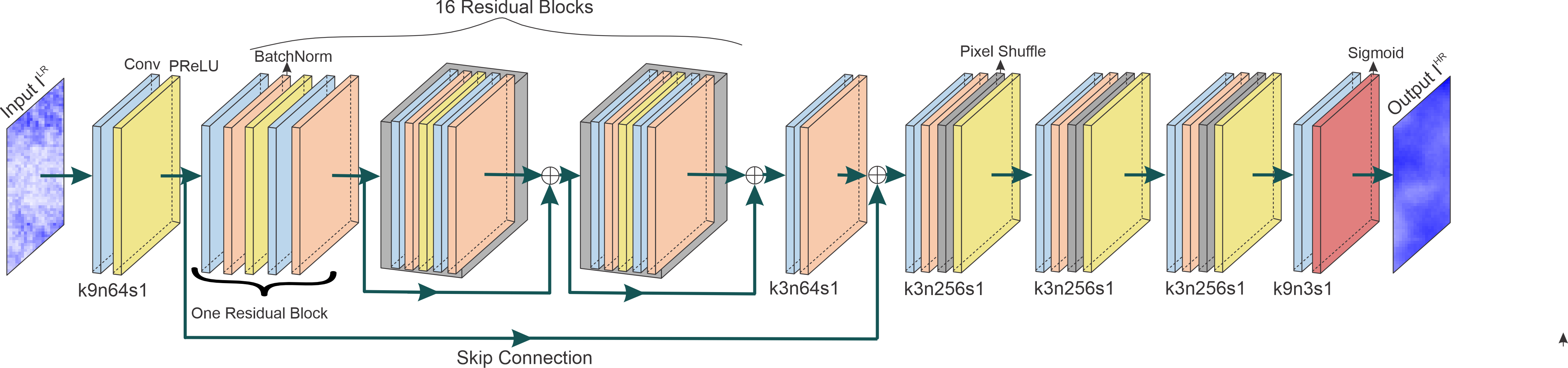}
        \caption{}
        \label{fig:sub1}
    \end{subfigure}
    
    \vspace{1em} % Add some space between the sub-figures
    
    \begin{subfigure}[b]{1\textwidth}
        \includegraphics[width=\textwidth]{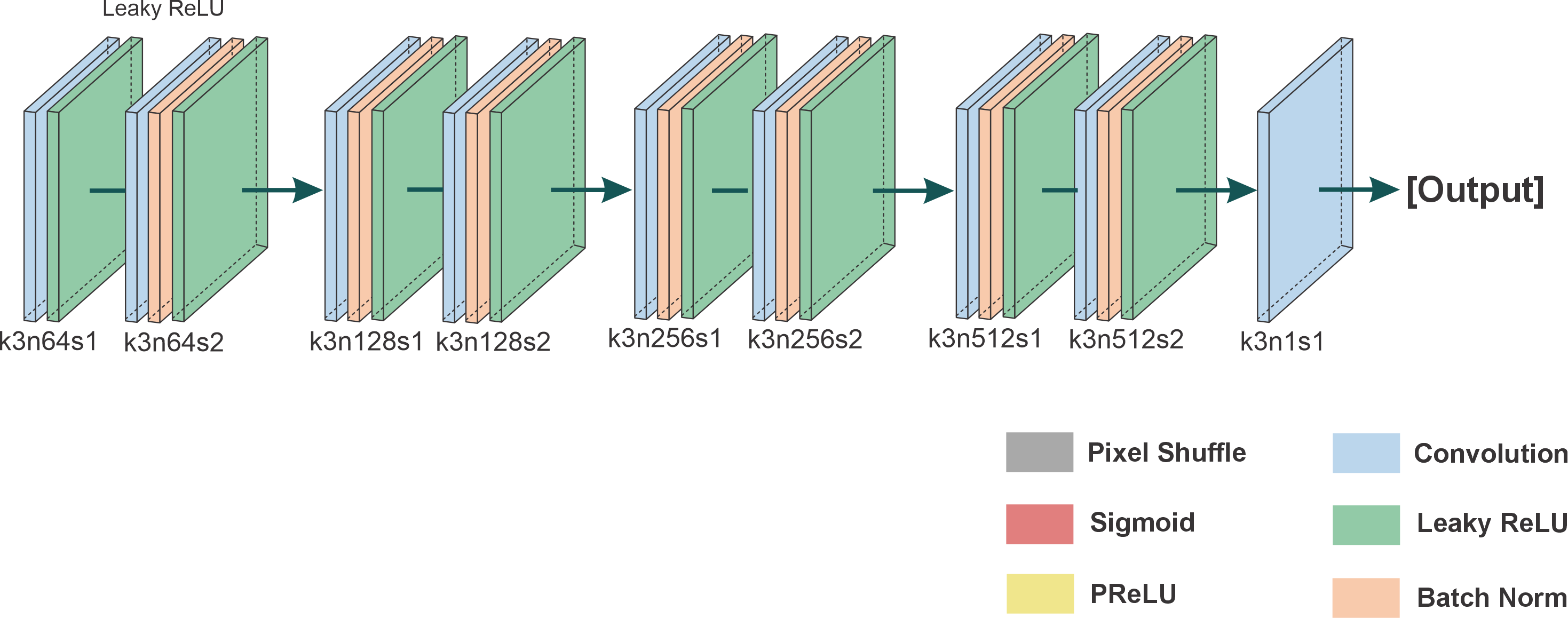}
        \caption{}
        \label{fig:sub2}
    \end{subfigure}
    
    \caption{Architecture of (a) generator, (b) discriminator}
    \label{fig:main}
\end{figure}

The SRGAN architecture used in this study incorporates a feature extractor derived from VGG-19, specifically leveraging its initial 18 layers. The generator's architecture is illustrated in Fig. \ref{fig:sub1}. The generator employs a ResNet architecture, which is known for its efficacy in dealing with vanishing and exploding gradient problems. The input goes through a convolutional layer with 64 filters, a kernel size of 9, stride of 1, and padding of 4, followed by a PReLU activation function. The output from the initial convolutional layer is passed through 16 residual blocks. Each residual block consists of two convolutional layers with 3x3 kernels, batch normalization, and PReLU activation. Later, the output is passed through another convolutional layer with 64 filters, 3x3 kernel size, stride of 1, and padding of 1, followed by batch normalization. The feature maps from the intermediate convolutional layer are then passed through three upsampling blocks. Each upsampling block consists of a 3x3 convolutional layer with 256 filters, batch normalization, pixel shuffle operation (upscale factor of 2), and a PReLU activation function. The output of the upsampling layers is passed through a final convolutional layer with three filters. The kernel size is 9x9, stride is 1, and padding is 4. The Sigmoid activation function is applied to ensure the output values are within the range [0, 1] as the RGB values are normalized to this range.

The discriminator as shown in Fig. \ref{fig:sub2} consists of four blocks. Each block contains two convolutional layers. The first convolutional layer performs a 3x3 convolution with a stride of 1 and padding of 1. If it's not the first block, batch normalization is applied after the first convolutional layer. Leaky ReLU activation with a negative slope of 0.2 follows each convolutional layer. The second convolutional layer performs a 3x3 convolution with a stride of 2 and padding of 1. Batch normalization is applied after the second convolutional layer. Another Leaky ReLU activation follows. After the four blocks, there is one more convolutional layer with a kernel size of 3x3, stride of 1, and padding of 1. This final convolutional layer reduces the number of channels to 1, effectively giving a single-channel output. No activation function is applied after this layer, making it a linear output.

In this study, we employ MSE loss for the discriminator, as opposed to binary cross-entropy. MSE loss measures the pixel-wise difference between generated and real images, aligning more closely with the purpose of our study which is increasing sensor data resolution. In addition, optimizing for pixel-wise similarity enhances stable training dynamics. SRGANs, notorious for their challenging training processes, benefit from MSE loss by mitigating issues like mode collapse and training instability. The smooth gradient landscape of MSE loss eases convergence for optimization algorithms, particularly in high-dimensional spaces prevalent in image generation tasks.

\section{Simulation Results}\label{results}
In this section, the simulation data and results are comprehensively detailed, showcasing the outcomes derived from the application of the described methodology to a specific and well-defined case study.

\subsection{Experimental Setup and Data}
The presented methodology was effectively utilized to analyze seismic data obtained from \cite{PEERDatabase}. This dataset comprises seismic data from 2014. The converted HR images are $136\times136$ while the LR images are assumed to be 64 times smaller, $17\times17$ in dimension. The parameters used for training the SRGAN model are given in Table \ref{params}.

\begin{table}[h!]
 	\centering
  %\vspace{-0.3cm}
 	\caption{SRGAN parameters}
        \vspace{-0.1cm}
 	\begin{tabular}{ll}
 	\hline
 	 	\textbf{Parameters} & \textbf{Values} \\
 		\hline
 		  learning rate & $0.0001$ \\
 		batch size & $32$\\
 		decay of first order momentum of gradient & $0.5$\\
            decay of second order momentum of gradient & $0.999$\\ 
            epoch to start learning rate decay & $250$\\
            adversarial loss coefficient $\lambda$ & $0.001$\\
            pixel loss coefficient $\beta$ & $10$\\
 		\hline\\
 	\end{tabular}%
  %\vspace{-0.3cm}
 	\label{params}%
 \end{table}%

%A comprehensive analysis of the data collected from the sensors is conducted aiming to gain valuable insights into the dataset.The data distribution for each sensor is visually represented in Fig.~\ref{}, providing a clear overview of the data patterns.

To assess the proposed method's efficacy, Structural Similarity Index (SSIM), Peak Signal-to-Noise Ratio (PSNR), and MSE are employed to compare the generated and real images. SSIM offers a perceptual evaluation, focusing on structural variances, brightness discrepancies, and textural diversities between images, aiming to align closer with human visual perception. According to \cite{Setiadi}, equations for these Indexes are as follow:

\begin{equation}
SSIM(i,i') = \frac{(2\mu_i\mu_{i'} + c_1)}{\mu_i^2 + \mu_{i'}^2 + c_1} \times \frac{(2\sigma_i\sigma_{i'} + c_2)}{\sigma_i^2 + \sigma_{i'}^2 + c_2} \times \frac{(\sigma_i{i'} + c_3)}{\sigma_i\sigma_{i'} + c_3}
\label{e5}
\end{equation}
where \( \mu_i \) and \( \mu_{i'} \) are the average pixel intensity of the subimages $i$ and $i'$. \( \sigma_i \), \( \sigma_{i'} \), and \( \sigma_{ii'} \) are the standard deviation for the $i$ and $i'$ subimages, and the covariance of the two subimages, respectively. Constant values $c_1$, $c_2$, and $c_3$ are used to avoid the zero denominators. MSE is calculated as

\begin{equation}
MSE = \frac{1}{M \times N \times O} \sum_{x=1}^{M} \sum_{y=1}^{N} \sum_{z=1}^{O} [I_{(x,y,z)} - I'_{(x,y,z)}]^2
\label{e6}
\end{equation}
where $M$ and $N$ are image resolution, $O$ is the number of image channels, $I_{(x,y,z)}$ is the pixel value of the original image at the $x$, $y$ coordinates and channel $z$, $I'$ is the output image processing result, in this research $I'$ is a stego image. PSNR is calculated as

\vspace{-0.5cm}

\begin{equation}
PSNR = 10 \log 10 \left(\frac{\text{max}^2}{MSE}\right)
\label{e2}
\end{equation}
Where $max$ is the highest scale value of the 8-bits grayscale.

\begin{figure}[b!]
    \centering
    \vspace{-0.3cm} 
    \begin{subfigure}[b]{0.5\textwidth}
        \includegraphics[width=1\textwidth]{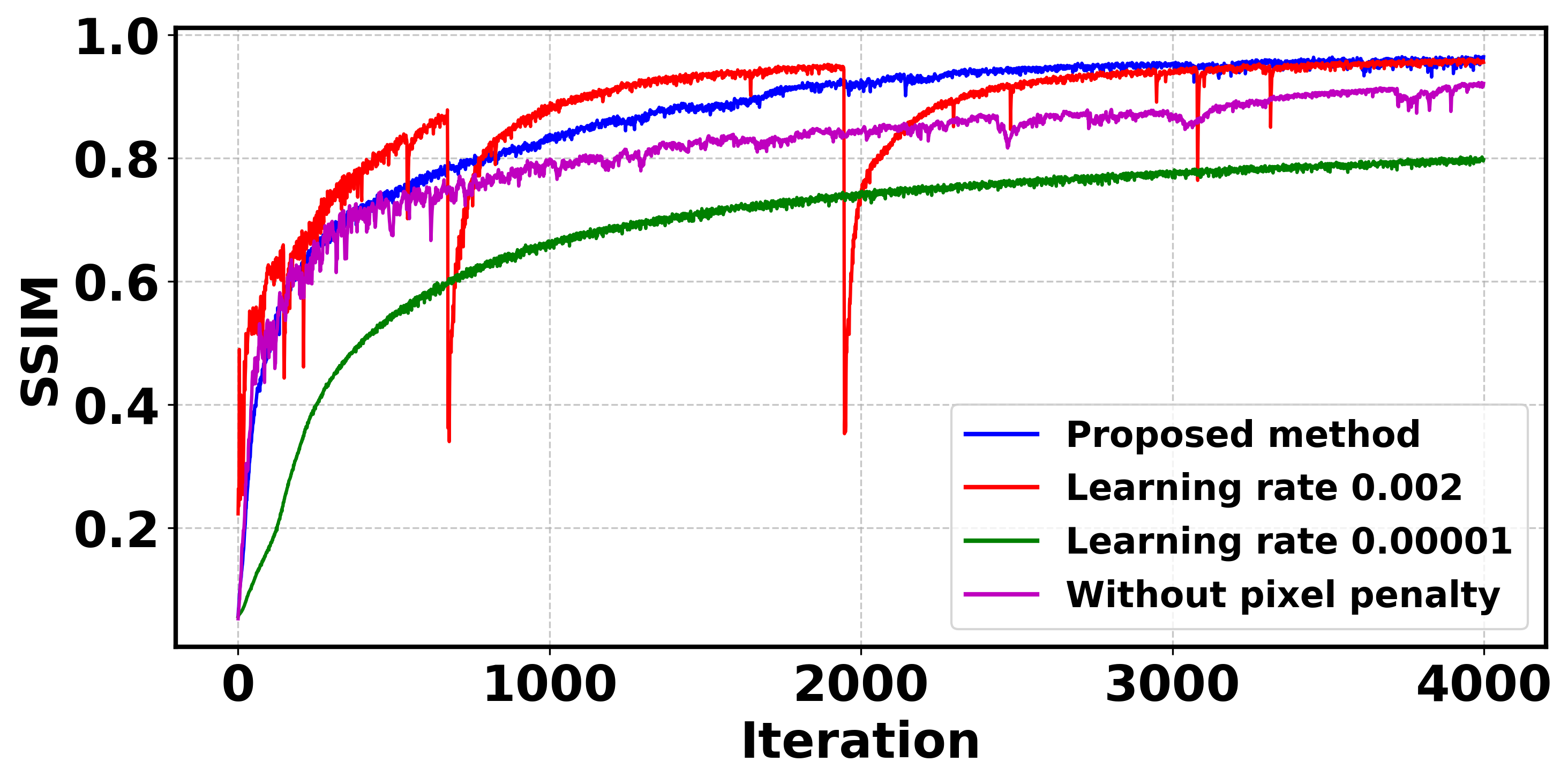}
        \caption{}
        \label{ssim}
    \end{subfigure}
    
    %\vspace{-0.1cm} % Add some space between the sub-figures
    
    \begin{subfigure}[b]{0.5\textwidth}
        \includegraphics[width=1\textwidth]{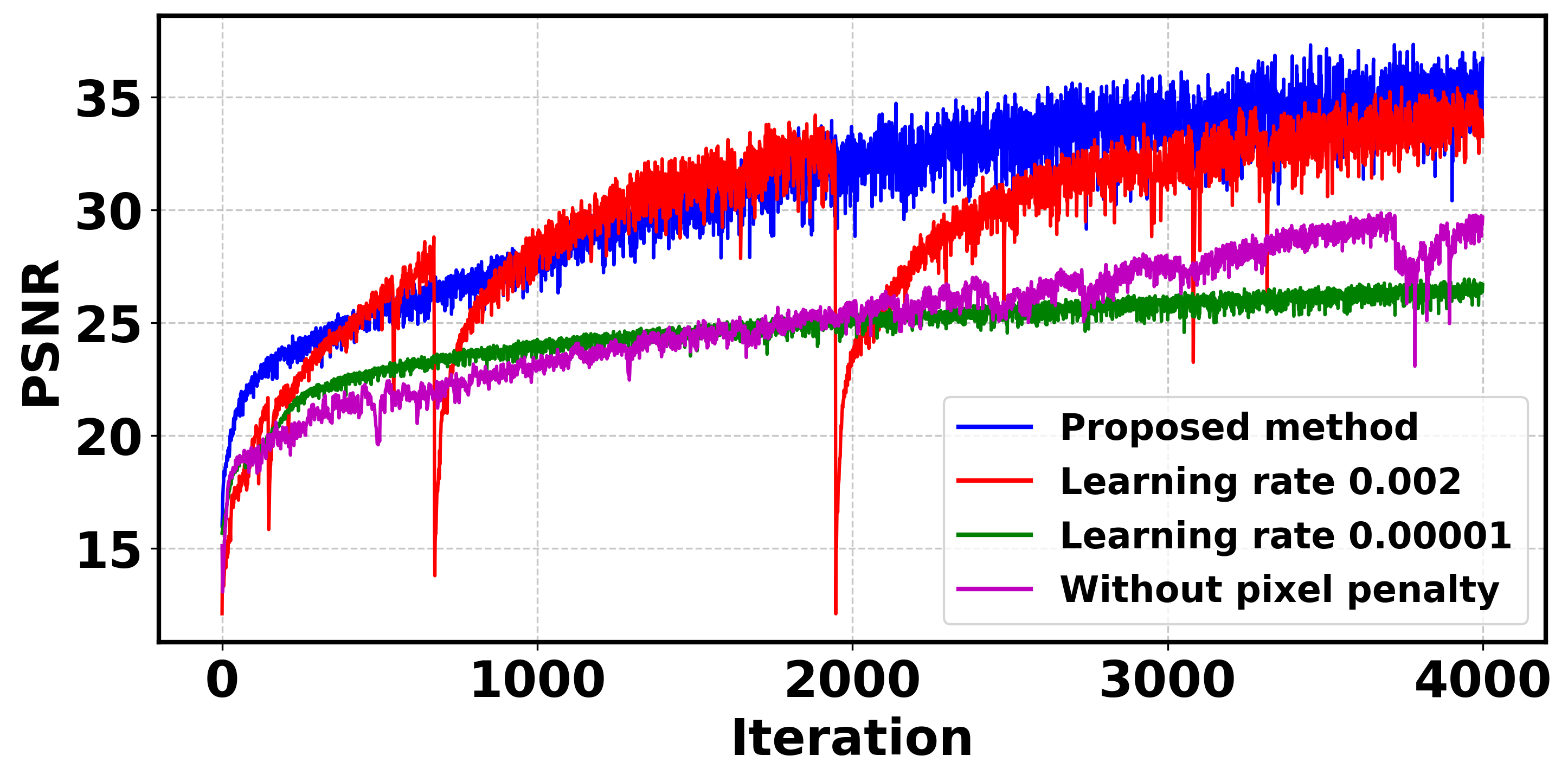}
        \caption{}
        \label{psnr}
    \end{subfigure}
    
    %\vspace{-0.1cm} 

    \begin{subfigure}[b]{0.5\textwidth}
        \includegraphics[width=1\textwidth]{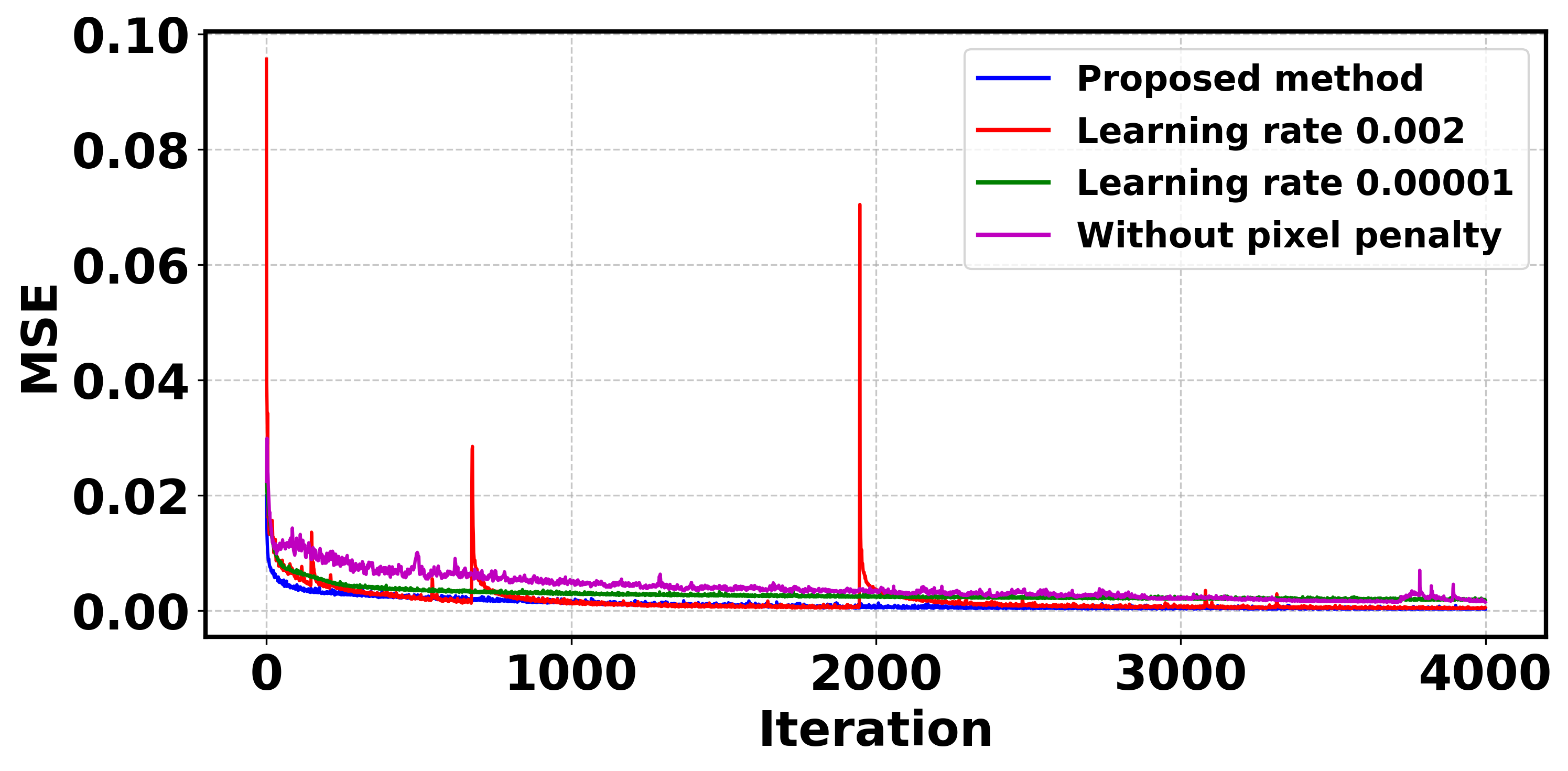}
        \caption{}
        \label{mse}
    \end{subfigure}
    \vspace{-0.3cm} 
    \caption{Performance metrics of the SRGAN model including (a) SSIM, (b) PSNR (c) MSE}
    \vspace{-0.3cm}
    \label{metrics}
\end{figure}

PSNR serves to quantify the disparity induced by noise or distortion, providing an insight into the visibility of elements such as compression anomalies or reconstruction inaccuracies in images. MSE, in contrast, provides a direct numeric representation of the aggregate squared deviations across corresponding pixels from the two images under comparison. The collective application of these metrics furnishes a multi-dimensional perspective, enabling a thorough evaluation of the proposed method's performance. Figure \ref{metrics} compares these metrics for the proposed method and three alternative structures of the model including: 1) model without the additional pixel penalty, 2) model with a higher learning rate of 0.002, and 3) model with a lower learning rate of 0.00001. Comparing the three figures reveals that the proposed method has performed proficiently, yielding high PSNR and SSIM values, and a low MSE, indicative of a minimal disparity between the original and generated images. These results collectively signify that the method has effectively increased image quality and structural integrity, confirming its viability and effectiveness.

In the process of training our SRGAN model, the generator's performance was monitored by evaluating its loss function, as depicted in Fig. \ref{loss} below. The generator's loss function is instrumental in guiding the network towards the generation of synthetic data that is indistinguishable from real data. Figure. \ref{loss} illustrates the trajectory of the generator’s loss over numerous training epochs. It can be observed from this figure that there is a notable decrease in the generator loss as the training progresses. This descending trend in the loss signifies that the generator is gradually improving in crafting data that more closely mimics the genuine data distribution. Such enhancement in performance is pivotal, as it allows the generator to produce more realistic and convincing synthetic data.

\begin{figure}[h!]
	\begin{center}
		\includegraphics[width=0.6\columnwidth]{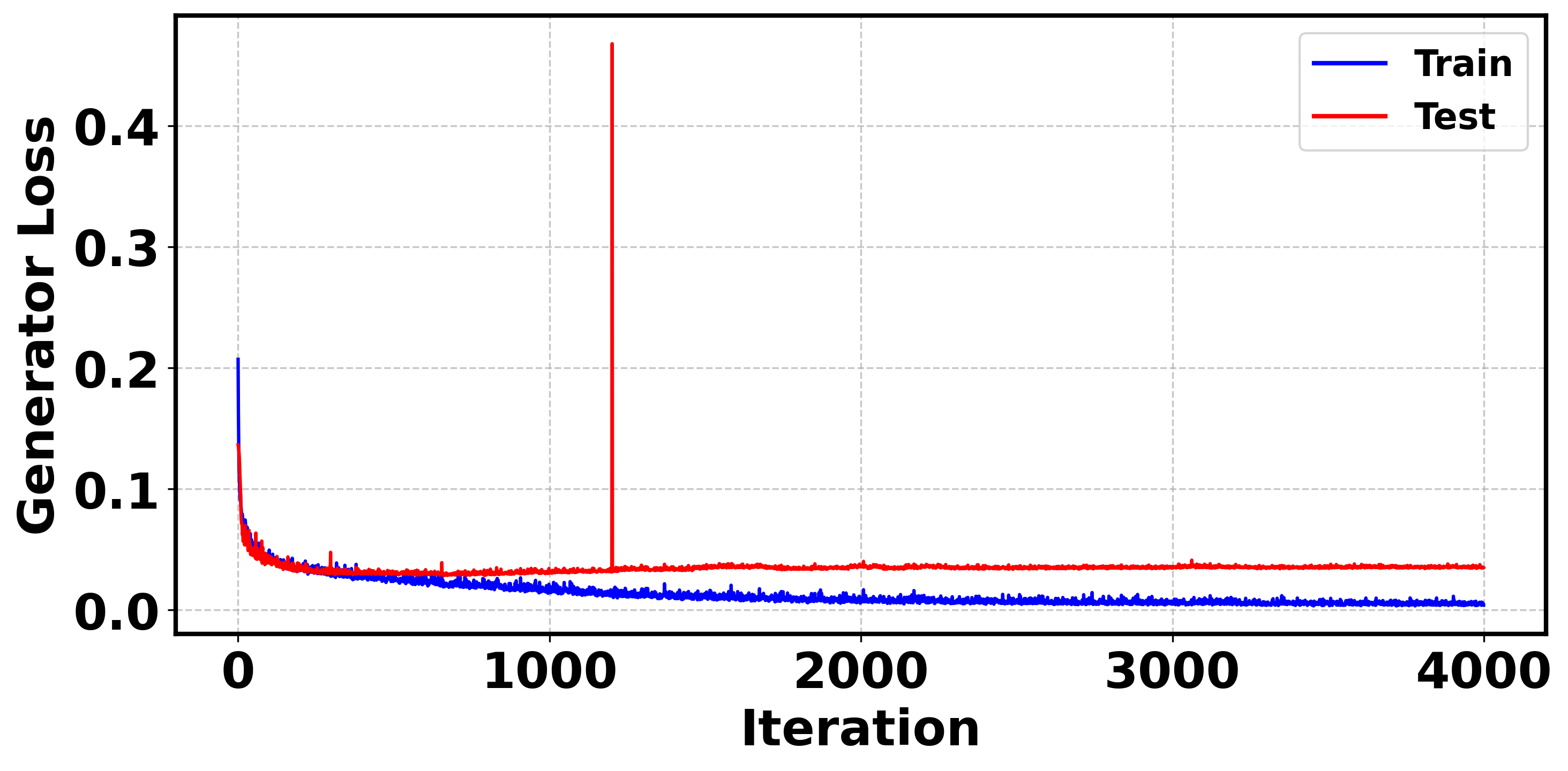}
		\caption{Generator loss for train and test data over training}\label{loss}
	\end{center}
\end{figure}

\begin{figure}[h!]
	\begin{center}
		\includegraphics[width=1\columnwidth]{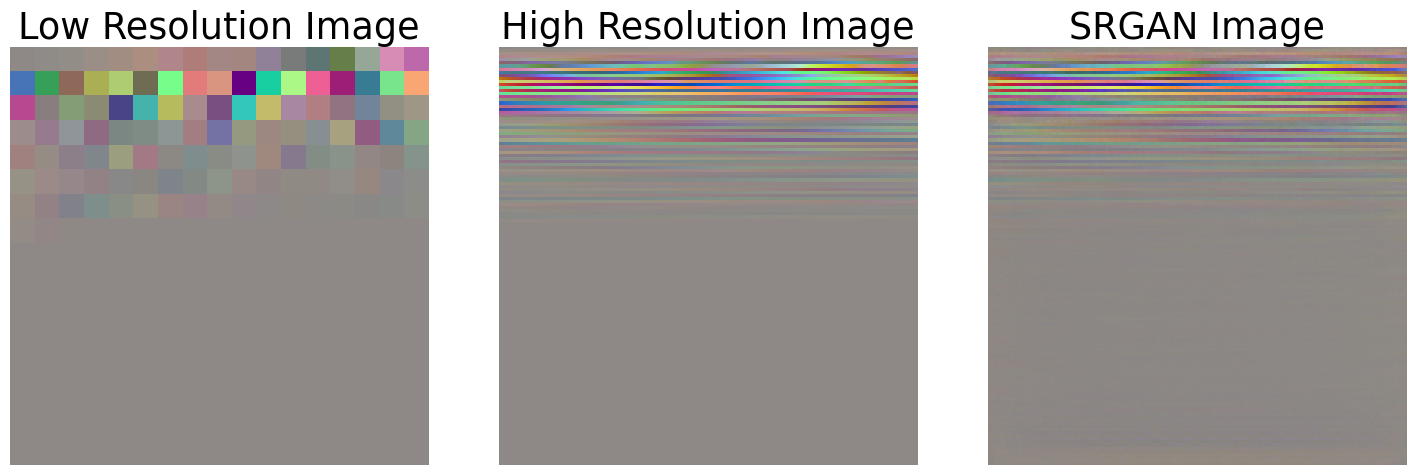}
		\caption{Comparison between an example low-resolution, high-resolution, and SRGAN-generated images from RSN752}\label{SRGAN_pic}
  \vspace{-0.5cm}
	\end{center}
\end{figure}

Figure \ref{SRGAN_pic} showcases three stages of image processing for a sample data. The leftmost image is in low-resolution, depicting the low-resolution sensor data. The center image is the actual high-resolution image produced from sensor data. On the right, the SRGAN-generated image is displayed, demonstrating a superior level of enhancement with remarkable detail and sharpness, representing the capabilities of advanced super-resolution techniques.

\begin{figure}[b!]
	\begin{center}
		\includegraphics[width=0.6\columnwidth]{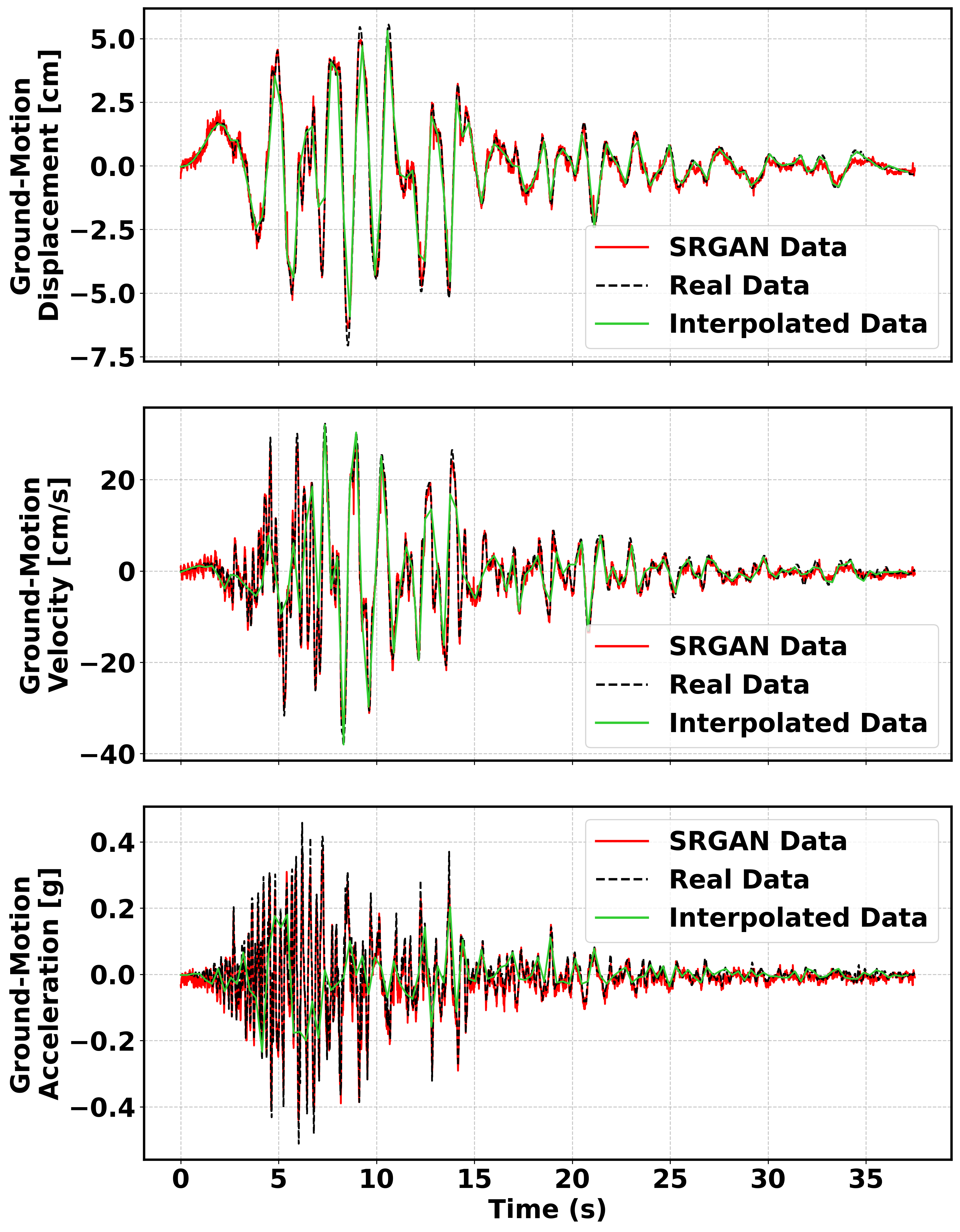}
		\caption{Comparative analysis of the transformed time series data from RSN752}\label{timeseries}
	\end{center}
\end{figure}

The generated images undergo a transformation back into time series data. In Fig. \ref{timeseries}, a comparison is conducted between the generated data, real data, and data obtained through a linear interpolation method for the same data shown in Fig. \ref{SRGAN_pic}. It is evident from this figure that the proposed SRGAN method is better in uncovering hidden structures, demonstrating enhanced performance compared to interpolation method. The MSE of the SRGAN method varies for different ground-motion measurements. Specifically, for ground-motion displacement, the MSE is 0.0446; for ground-motion velocity, it is 1.2862; and for ground-motion acceleration, it is 0.0005. In comparison, the interpolation method yields an MSE of 0.2546 for ground-motion displacement, 26.7901 for ground-motion velocity, and 0.0092 for ground-motion acceleration.

\begin{figure}[h!]
	\begin{center}
		\includegraphics[width=0.8\columnwidth]{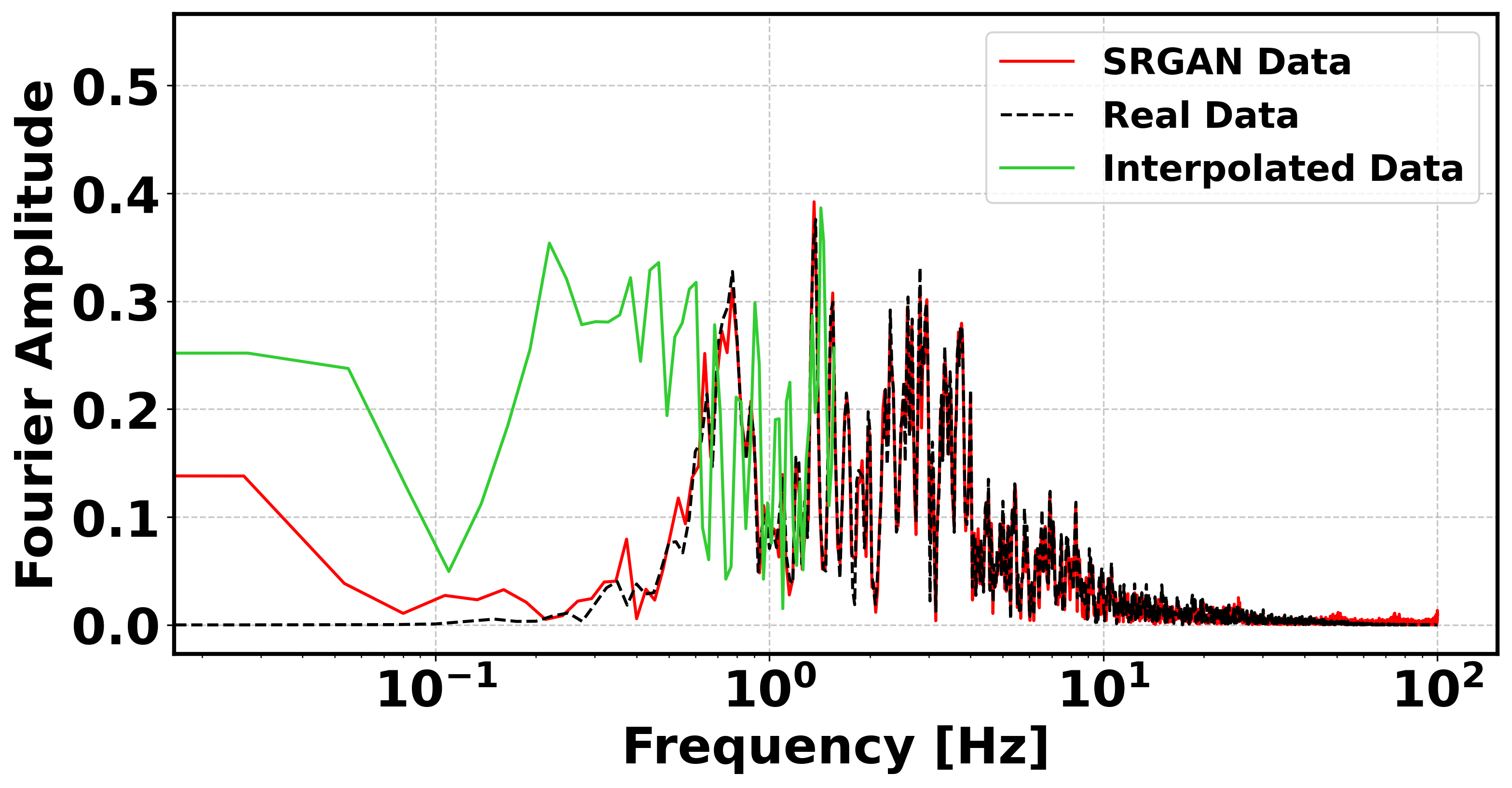}
		\caption{Comparison of Fourier amplitude spectrum of ground-motion acceleration for RSN752}\label{fft_}
	\end{center}
\end{figure}

It is believed that processing and interpretation of seismic data is more straight forward in frequency domain and using Fourier amplitude spectrum \cite{Fundamentals, Boore} and spatial interpolation of ground-motions have been completed in frequency domain in \cite{Thrainsson} and new studies use Fourier transform to better analyze and generate the ground-motion data \cite{Baglio}. Therefore, Fourier amplitude spectrum of acceleration is calculated in Fig. \ref{fft_} to reflect the efficiency of this method more properly. As can be seen from Fig. \ref{fft_}, most of signals from acceleration time-history are re-constructed. Therefore, this method can also be used to recover the signal transmission loss of using wireless sensors as described in \cite{Fan}.

\section{Conclusion}\label{Conclusion}
In this research, the novel application of SRGANs was explored to enhance the resolution of sensors in SHM systems, particularly in seismic-prone regions. By transforming sensor data into RGB images and subsequently using SRGANs to upscale these images, the study effectively addressed the challenges associated with high-resolution data acquisition, such as high costs and extensive storage needs. Comparative evaluations with conventional enhancement methods, using real seismic data, underscore the effectiveness of the proposed SRGAN technique. The research revealed that SRGAN significantly reduced the MSE for ground-motion displacement, ground-motion velocity, and ground-motion acceleration compared to the interpolation method. Specifically, it lowered the MSE values from 0.2546 to 0.0446 for displacement, from 26.7901 to 1.2862 for velocity, and from 0.0092 to 0.0005 for acceleration. The innovative approach not only simplifies the sensor network but also offers potential financial and storage efficiencies as it reduced the data size by 64 times. This advancement contributes a significant step towards achieving more sustainable and safer infrastructures globally, emphasizing the potential of SRGANs in improving SHM systems.

%Finally, authors believe that the accuracy of this study can be increased by selecting more appropriate data that have more common features by applying a clustering process before SRGAN training. Another recommendation is the selection of series of data with more correlation among them. This means we believe that SHM data of a special structure under a specific loading would have better results or in the case of our study, training of ground-motion data related to certain seismic fault type (i.e. strike-slip) with a specific range for closest distance to rupture plane ($R_{rup}$) or Joyner-Boore distance ($R_{JB}$) would improve the efficiency.

\bibliographystyle{elsarticle-num} 
\bibliography{References}

\begin{thebibliography}{10}
\expandafter\ifx\csname url\endcsname\relax
  \def\url#1{\texttt{#1}}\fi
\expandafter\ifx\csname urlprefix\endcsname\relax\def\urlprefix{URL }\fi
\expandafter\ifx\csname href\endcsname\relax
  \def\href#1#2{#2} \def\path#1{#1}\fi

\bibitem{Faroughi}
A.~Faroughi, M.~Hosseini, Simplification of earthquake accelerograms for quick time history analyses by using their modified inverse fourier transforms, Procedia Engineering 14 (2011) 2872--2877, the Proceedings of the Twelfth East Asia-Pacific Conference on Structural Engineering and Construction.
\newblock \href {https://doi.org/https://doi.org/10.1016/j.proeng.2011.07.361} {\path{doi:https://doi.org/10.1016/j.proeng.2011.07.361}}.

\bibitem{Phillips}
C.~Phillips, A.~R. Kottke, Y.~M. Hashash, E.~M. Rathje, Significance of ground motion time step in one dimensional site response analysis, Soil Dynamics and Earthquake Engineering 43 (2012) 202--217.
\newblock \href {https://doi.org/https://doi.org/10.1016/j.soildyn.2012.07.005} {\path{doi:https://doi.org/10.1016/j.soildyn.2012.07.005}}.

\bibitem{Mei}
Y.~Mei, D.~E. Hurtado, S.~Pant, A.~Aggarwal, On improving the numerical convergence of highly nonlinear elasticity problems, Computer Methods in Applied Mechanics and Engineering 337 (2018) 110--127.
\newblock \href {https://doi.org/https://doi.org/10.1016/j.cma.2018.03.033} {\path{doi:https://doi.org/10.1016/j.cma.2018.03.033}}.

\bibitem{resol1}
T.~Wu, G.~Liu, S.~Fu, F.~Xing, Recent progress of fiber-optic sensors for the structural health monitoring of civil infrastructure, Sensors 20~(16) (2020).
\newblock \href {https://doi.org/10.3390/s20164517} {\path{doi:10.3390/s20164517}}.

\bibitem{YU201760}
J.~Yu, P.~Zhu, B.~Xu, X.~Meng, Experimental assessment of high sampling-rate robotic total station for monitoring bridge dynamic responses, Measurement 104 (2017) 60--69.
\newblock \href {https://doi.org/https://doi.org/10.1016/j.measurement.2017.03.014} {\path{doi:https://doi.org/10.1016/j.measurement.2017.03.014}}.

\bibitem{naeim2008}
F.~Naeim, et~al., \href{https://www.latallbuildings.org/documents}{An Alternative Procedure For Seismic Analysis And Design Of Tall Buildings Located In The Los Angeles Region}, Los Angeles Tall Buildings Structural Design Council, 2020.
\newline\urlprefix\url{https://www.latallbuildings.org/documents}

\bibitem{Pacific}
F.~Naeim, et~al., \href{https://peer.berkeley.edu/research/building-systems/tall-buildings-initiative}{Guidelines for Performance-Based Seismic Design of Tall Buildings}, Pacific Earthquake Engineering Center, 2017.
\newline\urlprefix\url{https://peer.berkeley.edu/research/building-systems/tall-buildings-initiative}

\bibitem{resol2}
J.~Chakraborty, A.~Katunin, P.~Klikowicz, M.~Salamak, Early crack detection of reinforced concrete structure using embedded sensors, Sensors 19~(18) (2019).
\newblock \href {https://doi.org/10.3390/s19183879} {\path{doi:10.3390/s19183879}}.

\bibitem{resol3}
Z.~Lin, C.~Sun, W.~Liu, E.~Fan, G.~Zhang, X.~Tan, Z.~Shen, J.~Qiu, J.~Yang, A self-powered and high-frequency vibration sensor with layer-powder-layer structure for structural health monitoring, Nano Energy 90 (2021) 106366.
\newblock \href {https://doi.org/https://doi.org/10.1016/j.nanoen.2021.106366} {\path{doi:https://doi.org/10.1016/j.nanoen.2021.106366}}.

\bibitem{Pietro}
P.~D'Antuono, W.~Weijtjens, C.~Devriendt, On the minimum required sampling frequency for reliable fatigue lifetime estimation in structural health monitoring. how much is enough?, in: P.~Rizzo, A.~Milazzo (Eds.), European Workshop on Structural Health Monitoring, Springer International Publishing, Cham, 2023, pp. 133--142.

\bibitem{challenge1}
M.~A. Mousa, M.~M. Yussof, U.~J. Udi, F.~M. Nazri, M.~K. Kamarudin, G.~A.~R. Parke, L.~N. Assi, S.~A. Ghahari, Application of digital image correlation in structural health monitoring of bridge infrastructures: A review, Infrastructures 6~(12) (2021).
\newblock \href {https://doi.org/10.3390/infrastructures6120176} {\path{doi:10.3390/infrastructures6120176}}.

\bibitem{challenge3}
P.~Ragam, N.~Devidas~Sahebraoji, Application of {MEMS}-based accelerometer wireless sensor systems for monitoring of blast-induced ground vibration and structural health: A review, IET Wireless Sensor Systems 9~(3) (2019) 103--109.
\newblock \href {https://doi.org/https://doi.org/10.1049/iet-wss.2018.5099} {\path{doi:https://doi.org/10.1049/iet-wss.2018.5099}}.

\bibitem{challenge2}
C.~Niezrecki, J.~Baqersad, A.~Sabato, Digital Image Correlation Techniques for {NDE and SHM}, Springer International Publishing, Cham, 2018, pp. 1--46.
\newblock \href {https://doi.org/10.1007/978-3-319-30050-4_47-1} {\path{doi:10.1007/978-3-319-30050-4_47-1}}.

\bibitem{data}
A.~Fenerci, K.~A. Kvåle, Øyvind Wiig~Petersen, A.~Rønnquist, O.~Øiseth, Data set from long-term wind and acceleration monitoring of the {Hardanger} bridge, Journal of Structural Engineering 147~(5) (2021) 04721003.
\newblock \href {https://doi.org/10.1061/(ASCE)ST.1943-541X.0002997} {\path{doi:10.1061/(ASCE)ST.1943-541X.0002997}}.

\bibitem{Tang}
D.~Tang, J.~Chen, W.~Wu, L.~Jin, Q.~Yue, B.~Xie, S.~Wang, J.~Feng, Research on sampling rate selection of sensors in offshore platform {SHM} based on vibration, Applied Ocean Research 101 (2020) 102192.
\newblock \href {https://doi.org/https://doi.org/10.1016/j.apor.2020.102192} {\path{doi:https://doi.org/10.1016/j.apor.2020.102192}}.

\bibitem{Interpolation}
J.~Tong, X.~Chen, H.~Yu, Research on structural health monitoring method based on the inertial inclination sensor array, Journal of Physics: Conference Series 2264~(1) (2022) 012003.
\newblock \href {https://doi.org/10.1088/1742-6596/2264/1/012003} {\path{doi:10.1088/1742-6596/2264/1/012003}}.

\bibitem{signal_processing}
O.~S. Sonbul, M.~Rashid, Algorithms and techniques for the structural health monitoring of bridges: Systematic literature review, Sensors 23~(9) (2023).
\newblock \href {https://doi.org/10.3390/s23094230} {\path{doi:10.3390/s23094230}}.

\bibitem{GAN_medical1}
R.~Gupta, A.~Sharma, A.~Kumar, Super-resolution using {GANs} for medical imaging, Procedia Computer Science 173 (2020) 28--35, international Conference on Smart Sustainable Intelligent Computing and Applications under ICITETM2020.
\newblock \href {https://doi.org/https://doi.org/10.1016/j.procs.2020.06.005} {\path{doi:https://doi.org/10.1016/j.procs.2020.06.005}}.

\bibitem{GAN_medical2}
L.~Zhang, H.~Dai, Y.~Sang, A.~Bhardwaj, {Med-SRNet}: {GAN}-based medical image super-resolution via high-resolution representation learning 2022 (January 2022).
\newblock \href {https://doi.org/10.1155/2022/1744969} {\path{doi:10.1155/2022/1744969}}.

\bibitem{GAN_sattelite}
J.~Rabbi, N.~Ray, M.~Schubert, S.~Chowdhury, D.~Chao, Small-object detection in remote sensing images with end-to-end edge-enhanced {GAN} and object detector network, Remote Sensing 12~(9) (2020).
\newblock \href {https://doi.org/10.3390/rs12091432} {\path{doi:10.3390/rs12091432}}.

\bibitem{recovery}
Y.~J. Kim, D.~Hazra, Y.~Byun, K.~Ahn, Old document restoration using super resolution {GAN} and semantic image inpainting, in: Proceedings of the International Workshop on Artificial Intelligence and Education, WAIE 2019, Association for Computing Machinery, New York, NY, USA, 2020, p. 34–38.
\newblock \href {https://doi.org/10.1145/3397453.3397459} {\path{doi:10.1145/3397453.3397459}}.

\bibitem{PEERDatabase}
{Pacific Earthquake Engineering Research Center}, \href{https://ngawest2.berkeley.edu/}{Peer ground motion database} (accessed: September 16, 2023).
\newline\urlprefix\url{https://ngawest2.berkeley.edu/}

\bibitem{PEER201303}
T.~D. Ancheta, R.~B. Darragh, J.~P. Stewart, E.~Seyhan, W.~J. Silva, B.~S. Chiou, K.~E. Wooddell, R.~W. Graves, A.~R. Kottke, D.~M. Boore, T.~Kishida, J.~L. Donahue, {PEER NGA-West2} database, Tech. Rep. PEER Report 2013-03, Pacific Earthquake Engineering Research Center {(PEER)}, University of California, Berkeley, California, United States (2013).

\bibitem{fema}
{Applied Technology Council}, \href{https://nehrpsearch.nist.gov/article/PB2010-101512/XAB}{Qualification of building seismic performance factors}, Tech. Rep. FEMA-P695/ATC-63, Federal Emergency Management Agency, Washington, D.C., United States (2009).
\newline\urlprefix\url{https://nehrpsearch.nist.gov/article/PB2010-101512/XAB}

\bibitem{Setiadi}
D.~R. I.~M. Setiadi, {PSNR vs SSIM}: Imperceptibility quality assessment for image steganography, Multimedia Tools and Applications 80 (2021) 8423--8444.

\bibitem{Fundamentals}
Fundamentals of Signal Processing, 2012, pp. 25--158.
\newblock \href {http://arxiv.org/abs/https://library.seg.org/doi/pdf/10.1190/1.9781560801580.ch1} {\path{arXiv:https://library.seg.org/doi/pdf/10.1190/1.9781560801580.ch1}}, \href {https://doi.org/10.1190/1.9781560801580.ch1} {\path{doi:10.1190/1.9781560801580.ch1}}.

\bibitem{Boore}
D.~M. Boore, J.~J. Bommer, Processing of strong-motion accelerograms: Needs, options and consequences, Soil Dynamics and Earthquake Engineering 25~(2) (2005) 93--115.
\newblock \href {https://doi.org/https://doi.org/10.1016/j.soildyn.2004.10.007} {\path{doi:https://doi.org/10.1016/j.soildyn.2004.10.007}}.

\bibitem{Thrainsson}
H.~Thrainsson, A.~S. Kiremidjian, S.~R. Winterstein, \href{http://purl.stanford.edu/sg969zz1767}{Modeling of earthquake ground motion in the frequency domain}, John A Blume Earthquake Engineering Center Technical Report 134, Stanford Digital Repository (2000).
\newline\urlprefix\url{http://purl.stanford.edu/sg969zz1767}

\bibitem{Baglio}
M.~Baglio, A.~Cardoni, G.~P. Cimellaro, N.~Abrahamson, Generating ground motions using the fourier amplitude spectrum, Earthquake Engineering \& Structural Dynamics 52~(15) (2023) 4884--4899.
\newblock \href {http://arxiv.org/abs/https://onlinelibrary.wiley.com/doi/pdf/10.1002/eqe.3986} {\path{arXiv:https://onlinelibrary.wiley.com/doi/pdf/10.1002/eqe.3986}}, \href {https://doi.org/https://doi.org/10.1002/eqe.3986} {\path{doi:https://doi.org/10.1002/eqe.3986}}.

\bibitem{Fan}
G.~Fan, J.~Li, H.~Hao, Lost data recovery for structural health monitoring based on convolutional neural networks, Structural Control and Health Monitoring 26~(10) (2019) e2433, e2433 STC-19-0088.R1.
\newblock \href {http://arxiv.org/abs/https://onlinelibrary.wiley.com/doi/pdf/10.1002/stc.2433} {\path{arXiv:https://onlinelibrary.wiley.com/doi/pdf/10.1002/stc.2433}}, \href {https://doi.org/https://doi.org/10.1002/stc.2433} {\path{doi:https://doi.org/10.1002/stc.2433}}.

\end{thebibliography}

\end{document}